\documentclass{article}

\usepackage[numbers]{natbib}
\usepackage[preprint]{neurips_2025_custom}

\usepackage{amsmath,amsfonts,bm}

\def\eqref#1{equation~(\ref{#1})}
\def\Eqref#1{Equation~(\ref{#1})}

\def\1{\bm{1}}

\DeclareMathAlphabet{\mathsfit}{\encodingdefault}{\sfdefault}{m}{sl}
\SetMathAlphabet{\mathsfit}{bold}{\encodingdefault}{\sfdefault}{bx}{n}

\usepackage[dvipsnames]{xcolor}
\definecolor{linkColor}{rgb}{0.18,0.39,0.62}
\usepackage[utf8]{inputenc}
\usepackage[T1]{fontenc}
\usepackage[colorlinks=true,linkcolor=linkColor,citecolor=linkColor,filecolor=linkColor,urlcolor=linkColor]{hyperref}
\usepackage{cleveref}
\usepackage{multirow}
\usepackage{xspace}
\usepackage{booktabs}
\usepackage{graphicx}
\usepackage{placeins}
\usepackage{array}
\usepackage{wrapfig}
\usepackage{subcaption}
\usepackage{bm}
\usepackage{algorithm}

\usepackage{algpseudocode}
\usepackage{enumitem}
\usepackage{tcolorbox}
\usepackage{makecell}
\usepackage{diagbox}

\usepackage{amsmath}
\usepackage{amssymb}
\usepackage{mathtools}
\usepackage{amsthm}

\usepackage{multirow}
\usepackage{amsmath}
\usepackage{capt-of}
\usepackage{tabularx}
\usepackage{epsfig}
\usepackage{amssymb}
\usepackage{amsfonts}
\usepackage{booktabs}
\usepackage{scalerel}
\usepackage{listings}
\usepackage{varwidth}
\usepackage{stmaryrd}
\usepackage{bbm}
\usepackage{wrapfig}
\usepackage{pifont}

\newcommand{\tabincell}[2]{\begin{tabular}{@{}#1@{}}#2\end{tabular}}

\definecolor{deepblue}{rgb}{0,0,0.5}
\definecolor{officeblue}{RGB}{0,102,204}
\definecolor{deepred}{rgb}{0.6,0,0}
\definecolor{deepgreen}{rgb}{0,0.5,0}
\definecolor{mybrickred}{RGB}{182,50,28}

\definecolor{fillcolor}{RGB}{216,217,252}

\usepackage{etoolbox}
\usepackage{framed}

\newif\ifxetexorluatex
\ifxetex
  \xetexorluatextrue
\else
  \ifluatex
    \xetexorluatextrue
  \else
    \xetexorluatexfalse
  \fi
\fi
\newcommand*\quotesize{60} 
\newcommand*{\openquote}
   {\tikz[remember picture,overlay,xshift=-4ex,yshift=-2.5ex]
   \node (OQ) {\fontsize{\quotesize}{\quotesize}\selectfont``};\kern0pt}

\newcommand*{\closequote}[1]
  {\tikz[remember picture,overlay,xshift=4ex,yshift={#1}]
   \node (CQ) {\fontsize{\quotesize}{\quotesize}\selectfont''};}

\colorlet{shadecolor}{white}

\newcommand*\shadedauthorformat{\emph} 

\newcommand*\authoralign[1]{%
  \if#1l
    \def\authorfill{}\def\quotefill{\hfill}
  \else
    \if#1r
      \def\authorfill{\hfill}\def\quotefill{}
    \else
      \if#1c
        \gdef\authorfill{\hfill}\def\quotefill{\hfill}
      \else\typeout{Invalid option}
      \fi
    \fi
  \fi}
{\authoralign{#1}
\ifblank{#2}
   {\def\shadequoteauthor{}\def\yshift{-2ex}\def\quotefill{\hfill}}
   {\def\shadequoteauthor{\par\authorfill\shadedauthorformat{#2}}\def\yshift{2ex}}
\begin{snugshade}\begin{quote}\openquote}
{\shadequoteauthor\quotefill\closequote{\yshift}\end{quote}\end{snugshade}}

\newcommand{\github}{\raisebox{-1.5pt}{\includegraphics[height=1.05em]{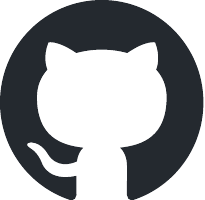}}\xspace}

\definecolor{DarkBlue}{RGB}{0, 51, 153}

\title{Learning to Coach for Experiential Learning}

\author{%
Guanheng Chen$^{1,2}$\thanks{~Equal contribution.}~~~~~~~~Tianzhu Ye$^{1}$\footnotemark[1]~~~~~~~~Li Dong$^{1}$\footnotemark[1] \\
\bf Xun Wu$^{1}$~~~~~~Shaohan Huang$^{1}$~~~~~~Furu Wei$^{1}$ \\
$^{1}$~Microsoft Research \\
$^{2}$~Tsinghua University \\
~{\href{https://aka.ms/GeneralAI}{https://aka.ms/GeneralAI}}
}

\begin{document}

\maketitle

\begin{abstract}
Language models can learn from experience, but raw solution trajectories
are often too long and noisy to provide effective guidance. In this work, we
propose \textbf{Learning to Coach (L2C)}, a framework that trains a dedicated
LLM-as-a-Coach to extract actionable experiential knowledge from an actor
model's previous trajectory. The actor remains frozen, while the LLM-as-a-Coach is
trained to maximize a reward given by the correctness of the actor's
guided response. We study two such rewards: a same-instance reward, which
improves subsequent responses on the original problem, and a cross-instance
reward, which elicits knowledge that transfers to other instances. Across
mathematical reasoning and interactive text-games, L2C consistently outperforms
self-refinement and an untrained LLM-as-a-Coach. Running experiential learning
for more iterations further improves accuracy and uses additional inference
compute more effectively than enlarging the actor's decoding budget. The trained
LLM-as-a-Coach also transfers to out-of-distribution tasks and adapts its
guidance to the specific actor it coaches.
\begin{table}[H]
\centering
\begin{tabular}{@{}r@{\hspace{2pt}}l@{}}
\github & \textbf{Code}: \href{https://aka.ms/l2c-code}{\texttt{aka.ms/l2c-code}}
\end{tabular}
\end{table}
\end{abstract}

\vfill{}

\begin{figure}[h]
\centering
\includegraphics[width=\linewidth]{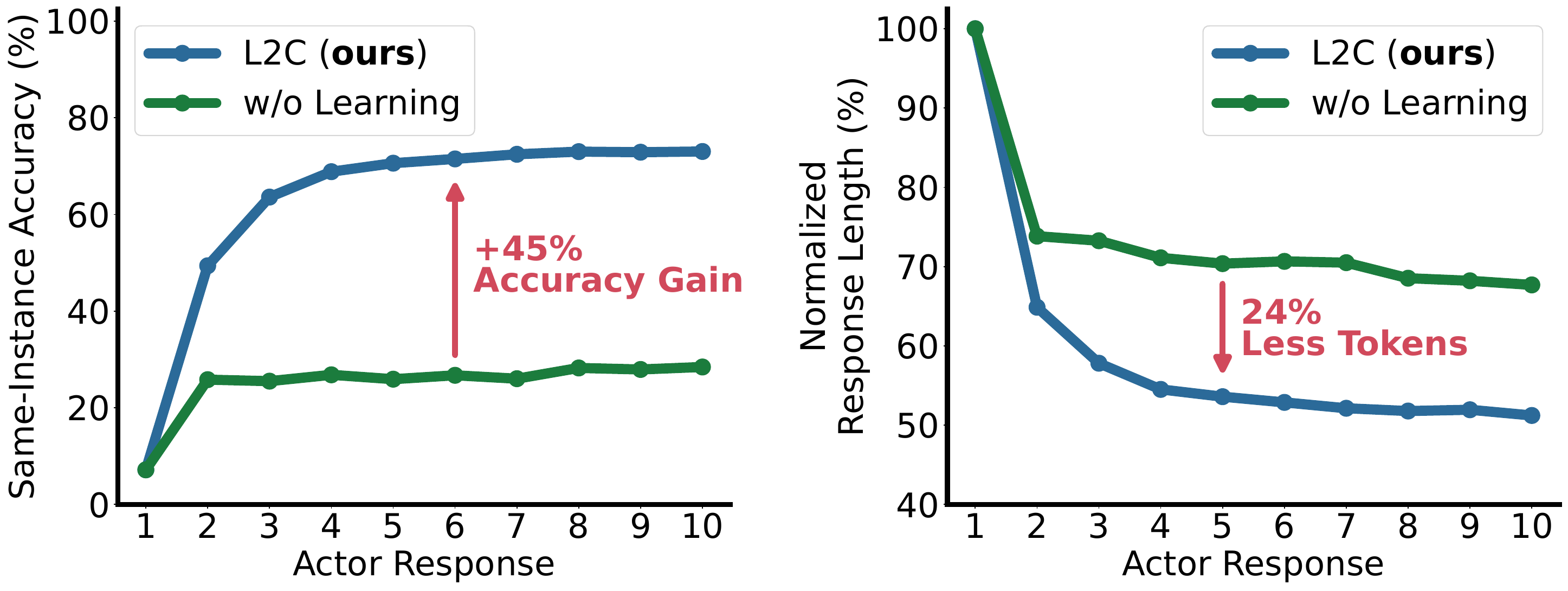}
\caption{Through experiential learning, a trained LLM-as-a-Coach steadily improves
the actor's accuracy, while an untrained LLM-as-a-Coach quickly plateaus (left).
Meanwhile, responses become progressively shorter (right), indicating that learned coaching enables more accurate and efficient test-time scaling.}
\end{figure}

\vfill{}

\newpage
\section{Introduction}
\label{sec:intro}
Large language models (LLMs) have demonstrated strong capabilities in
mathematical reasoning~\citep{deepseekr1,qwen3} and, as language agents, in
planning and interactive decision making~\citep{voyager,generativeagents}.
Beyond solving a problem in a single pass, an LLM can
often improve its answer by inspecting an earlier response, identifying errors,
and trying again. Its own interaction history therefore carries valuable
\emph{experiential knowledge}: failed approaches, useful intermediate results,
latent environment rules, and promising directions for subsequent responses.

However, extracting useful knowledge from a solving trajectory is itself a
challenging reasoning problem. A raw trajectory can be long, redundant, or
misleading, and may contain incorrect conclusions alongside useful evidence.
Simply conditioning the model on its entire previous response, as in
self-refinement, does not distinguish between these components
\citep{selfrefine,rl4f}. As a result, the model may repeat its original
mistakes, attend to irrelevant details, or even regress after an additional
solve. Alternatively, one can fine-tune the solving model directly using
task-level rewards, but doing so changes the model's parameters and can be
expensive when the same model must retain broad, general-purpose
capabilities.

This motivates a different division of labor: instead of updating the model
that performs the task, we train a separate model to learn how to extract
useful experience from prior trajectories, while keeping the task-performing model
frozen. This separation lets the solving model retain its original
parameters while delegating the interpretation of prior trajectories to a
dedicated model. It also makes the extracted knowledge explicit and inspectable, rather than
implicitly encoded.

We term this delegated model an \emph{LLM-as-a-Coach}, as in
\citep{coachel}. An
LLM-as-a-Coach reads the actor's prior trajectory and expresses, in natural language, what
went wrong, which intermediate results are worth keeping, and which directions
are promising for the next response. It distills the trajectory into concise, actionable experiential knowledge
for the actor's subsequent response.

In this work, we propose \textbf{Learning to Coach (L2C)}, a framework that
trains a dedicated LLM-as-a-Coach to extract actionable experiential knowledge
from a frozen actor model's previous solving trajectory. In L2C, experiential
learning proceeds as follows: the actor first makes an initial response, the
LLM-as-a-Coach converts the resulting trajectory into concise experiential knowledge,
and the actor makes a guided response. Rather
than supervising the LLM-as-a-Coach with human-written critiques, we optimize it with
reinforcement learning, using the correctness of the actor's guided response as the
reward. The LLM-as-a-Coach therefore learns to produce guidance that improves the
downstream behavior of the specific actor being coached.

We study two reward variants. The \emph{same-instance} reward evaluates the
guided response on the instance from which the experiential knowledge was
extracted, encouraging the LLM-as-a-Coach to diagnose errors and preserve useful
intermediate results. The \emph{cross-instance} reward evaluates the extracted
knowledge on disjoint instances, encouraging it to capture reusable task
structure. L2C also supports experiential learning over multiple iterations, where
the LLM-as-a-Coach repeatedly updates its experiential knowledge from the actor's
latest trajectory before the actor solves again.

We evaluate L2C across two domains, mathematical reasoning and interactive
text-games. Across tasks and model sizes, L2C consistently
outperforms the frozen base actor, Self-Refinement, and an untrained
LLM-as-a-Coach. The gains are particularly large in text-games, where
interaction trajectories expose latent environment rules. Training with the
cross-instance reward further produces transferable knowledge when such rules
are shared across instances.

Our analysis shows that experiential learning provides an effective form of
test-time scaling: performance continues to improve across iterations,
while an untrained LLM-as-a-Coach quickly plateaus. On math, L2C with more
iterations outperforms doubling the actor's decoding budget across all evaluated
model sizes. The
trained LLM-as-a-Coach also transfers to out-of-distribution
evaluations, and adapts its guidance to the specific actor. These results show that
learning to extract actor-specific experiential knowledge can improve language
models without updating their parameters.


\begin{figure}[t]
\centering
\begin{tcolorbox}[width=\linewidth, colback=white, colframe=white,
                  boxrule=0pt, arc=0pt, left=5pt, right=5pt, top=4pt, bottom=4pt]
\footnotesize
\newcommand{\flcell}[1]{\makebox[0.85em][c]{\rule[-0.25em]{0pt}{1.05em}#1}}
\newcommand{\flboard}[9]{%
  {\setlength{\tabcolsep}{1.5pt}\renewcommand{\arraystretch}{1}\ttfamily\scriptsize
   \begin{tabular}{|c|c|c|}\hline
   \flcell{#1} & \flcell{#2} & \flcell{#3} \\\hline
   \flcell{#4} & \flcell{#5} & \flcell{#6} \\\hline
   \flcell{#7} & \flcell{#8} & \flcell{#9} \\\hline\end{tabular}}}
\begin{minipage}[c]{0.14\linewidth}\centering
\textbf{Source Problem}\\[3pt]
\flboard{P}{~}{H}{H}{~}{~}{~}{~}{G}\\[3pt]
{\scriptsize One episode by the frozen actor.}
\end{minipage}
\hfill
\begin{minipage}[c]{0.175\linewidth}\centering
\fcolorbox{black!45}{black!6}{\shortstack{%
  {\scriptsize\bfseries LLM-as-a-Coach}\\[1pt]
  {\scriptsize $\pi_\theta$}}}\\[3pt]
$\xRightarrow{\hspace{3.2em}}$\\[-4pt]
{\scriptsize Extract}
\end{minipage}
\hfill
\begin{minipage}[c]{0.26\linewidth}
{\centering\textbf{Experiential knowledge}\par}
\vspace{3pt}
\scriptsize
The game involves navigating a 3x3 grid where the \textcolor{ForestGreen}{player (P) starts} and
must reach \textcolor{ForestGreen}{the goal (G)}. The grid contains \textcolor{ForestGreen}{holes (H) that
end the episode}. The player can move up, down, left, or right, and
\textcolor{ForestGreen}{must avoid holes while reaching the goal}.
\end{minipage}
\hfill\hspace{6pt}$\xRightarrow{\raisebox{3pt}{\scriptsize Guided Response}}$\hfill
\begin{minipage}[c]{0.24\linewidth}\centering
\textbf{Target Problem}\\[1pt]
{\scriptsize
\textbf{Same problem}\\[-1pt]
same-instance reward ($r_{\mathrm{same}}$)\\[1pt]
\textbf{Unseen problem}\\[-1pt]
cross-instance reward ($r_{\mathrm{cross}}$)}\\[3pt]
\flboard{P}{~}{~}{~}{~}{~}{H}{H}{G}\\[3pt]
{\scriptsize\raggedright
\textcolor{BrickRed}{$\times$}~Fell into the hole, reward $0$\\
\textcolor{ForestGreen}{$\checkmark$}~Goal reached, reward $1$\par}
\end{minipage}
\end{tcolorbox}
\vspace{0.5cm}

\includegraphics[width=0.8\linewidth]{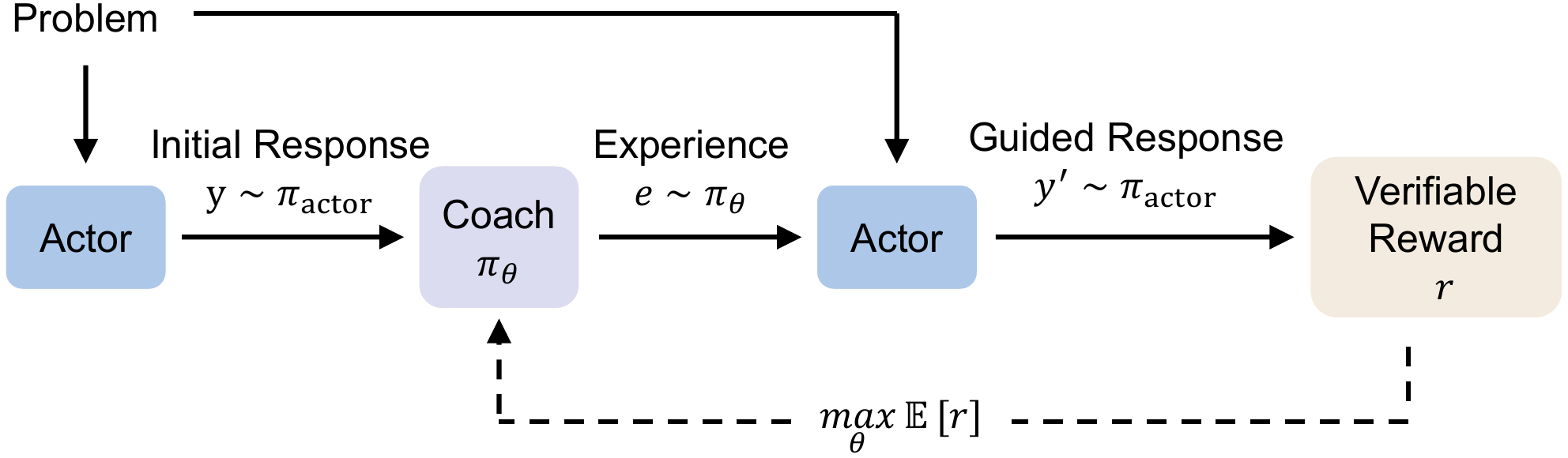}
\vspace{-0.1cm}
\caption{Overview of L2C. A frozen actor $\pi_{\mathrm{actor}}$ first responds to
problems, producing trajectories; an LLM-as-a-Coach $\pi_\theta$ extracts
experiential knowledge from each trajectory. The actor then produces a guided
response conditioned on the extracted knowledge, and the verifiable correctness
of that response serves as the reward. Only the
LLM-as-a-Coach is trained. Under the same-instance reward, the target is the
source problem; under the cross-instance reward, the target is an unseen
problem.}
\vspace{-0.1cm}
\label{fig:method}
\end{figure}

\section{Method}
\label{sec:method}

We propose \textbf{Learning to Coach (L2C)}, which learns from trajectories
produced by a frozen actor during deployment. Once these trajectories have been
collected, an LLM-as-a-Coach extracts concise experiential
knowledge from the given trajectory, which is supplied back to the actor to improve its later performance.
Unlike conventional fine-tuning, which modifies the actor directly, L2C keeps the
actor fixed and instead optimizes the coach to produce experiential knowledge
that improves the actor's performance.
\paragraph{Learning to Coach}
Let $\pi_{\mathrm{actor}}$ be a frozen actor and $\pi_\theta$ the trainable
LLM-as-a-Coach. Running the actor on an instance $x$ leaves behind a trajectory
$(x,y)$, where $y \sim \pi_{\mathrm{actor}}(\cdot\mid x)$, which may be a solution trace or a
multi-turn interaction gathered during actor use. Let $\mathcal{D}$ denote the resulting
distribution over collected trajectories. Given $(x,y) \sim \mathcal{D}$, the coach
extracts experiential knowledge
\begin{equation}
e \sim \pi_\theta(\cdot\mid x,y).
\end{equation}
To reward the experiential knowledge $e$, we provide it to the frozen actor on a target instance
$z$:
\begin{equation}
y'_z \sim \pi_{\mathrm{actor}}(\cdot\mid z,e),
\qquad
r(z,e)=\mathcal{V}(z,y'_z),
\end{equation}
where $\mathcal{V}$ is a deterministic verifier. Thus, the coach is
rewarded when its guidance leads the actor to a correct response.
Accordingly, we train the coach to maximize this reward:
\begin{equation}
\max_{\theta} \;\; \mathbb{E}_{\, (x,y) \sim \mathcal{D},\; e} \left[\, r \,\right],
\label{eq:objective}
\end{equation}
where $r$ is one of two reward variants, which differ only in which target instances the experiential knowledge is evaluated
on.

The \emph{same-instance reward} evaluates the knowledge on the instance $x$
that $y$ was collected on,
\begin{equation}
r_{\mathrm{same}} = r(x,e),
\label{eq:same-reward}
\end{equation}
which encourages the coach to distill instance-specific knowledge. 

The \emph{cross-instance reward} instead evaluates the knowledge on a disjoint set
of target instances $\mathcal{P}$,
\begin{equation}
r_{\mathrm{cross}}
=
\frac{1}{|\mathcal{P}|}\sum_{z\in\mathcal{P}}r(z,e),
\label{eq:meta-reward}
\end{equation}
which encourages reusable knowledge, such as shared task rules and general
strategies.

For each source trajectory, we sample multiple candidate knowledge snippets,
evaluate each through actor rollouts, and optimize the coach with
GRPO~\citep{grpo}.

\paragraph{Experiential Learning}
LLM-as-a-Coach can repeatedly update its experiential knowledge as additional actor trajectories
become available. Starting with $e^{(0)}=\emptyset$, iteration $k$ performs
\begin{align}
y^{(k)}
&\sim \pi_{\mathrm{actor}}(\cdot\mid x,e^{(k-1)}),\\
e^{(k)}
&\sim \pi_\theta(\cdot\mid x,y^{(k)},e^{(k-1)}).
\end{align}
Here $K$ denotes the total number of actor responses, counting the initial
unguided response $y^{(1)}$. The updated knowledge replaces the previous version, allowing the coach to
remove incorrect conclusions, retain useful discoveries, and refine its
guidance over time. The $K{=}2$ setting uses one observed trajectory followed
by one guided response; larger $K$ models continued learning from accumulated
deployment experience. Viewed from a meta-learning perspective~\citep{maml}, generating the
experience from new actor trajectories forms the inner loop, while updating the
coach with reinforcement learning based on the actor's downstream reward forms
the outer loop.
Algorithm~\ref{alg:l2c} summarizes the training procedure.

\begin{algorithm}[t]
\small
\caption{Experiential Learning with L2C}
\label{alg:l2c}
\begin{algorithmic}
\Require Collected trajectories $\mathcal{D}$; Frozen actor
$\pi_{\mathrm{actor}}$; LLM-as-a-Coach $\pi_\theta$; Iterations $K$;
Reward $r \in \{r_{\mathrm{same}}, r_{\mathrm{cross}}\}$
\Ensure Trained LLM-as-a-Coach $\pi_\theta$
\For{each collected trajectory $(x, y^{(1)}) \sim \mathcal{D}$}
    \State $e^{(0)} \gets \emptyset$
    \For{$k \gets 1$ \textbf{to} $K-1$}
        \State $e^{(k)} \sim \pi_\theta(\cdot \mid x, y^{(k)}, e^{(k-1)})$
        \Comment{Extract experiential knowledge}
        \State $y^{(k+1)} \sim
        \pi_{\mathrm{actor}}(\cdot \mid x, e^{(k)})$
        \Comment{Guided response}
        \State Update $\pi_\theta$ to maximize $\mathbb{E}[r]$
        \Comment{\Eqref{eq:objective}}
    \EndFor
\EndFor
\State \Return $\pi_\theta$
\end{algorithmic}
\end{algorithm}

\section{Experiments}
\label{sec:experiments}

\subsection{Setup}
\label{sec:setup}

\paragraph{Base Models}
In every setting, the actor and the LLM-as-a-Coach are initialized from the same
base model. For math, we use Qwen3-1.7B, Qwen3-4B, and Qwen3-8B~\citep{qwen3},
all in thinking mode. For text-games, we use Qwen3-1.7B (thinking mode) on
FrozenLake-v0-raw; on Sokoban-v0 we use Qwen3-4B (thinking mode) in the $K{=}2$
setting and Qwen3-4B-Instruct-2507 in the $K{=}10$ setting.  

\paragraph{Datasets}
Our experiments cover three datasets: the math corpus
DAPO-Math-17K~\citep{dapo} and two interactive text-game environments,
FrozenLake and Sokoban, both built on TextArena~\citep{textarena}.
DAPO-Math-17K comprises roughly $14$K English math problems with verifiable
numerical answers. FrozenLake asks the agent to navigate a grid to a goal
tile while steering clear of holes, whereas Sokoban requires planning a
sequence of pushes that moves a box onto a target without dropping into a
hole or jamming the box against a wall. Both games are presented purely as
text and played out over multiple interaction turns. Full per-dataset
configurations are deferred to Appendix~\ref{app:exp_detail_dataset}.

\paragraph{Training}
We optimize the LLM-as-a-Coach with GRPO~\citep{grpo} while keeping the actor
frozen. For each source trajectory, we sample $n{=}8$ candidate
experiential-knowledge snippets from the LLM-as-a-Coach; their rewards form one
GRPO advantage-normalization group. Each candidate conditions an independent
guided response from the frozen actor, which the verifier $\mathcal{V}$ scores
as a binary reward: answer correctness on math, and the outcome of the guided
playthrough on text-games. Actor responses are limited to $16{,}384$ tokens on
math and, on text-games, to five interaction turns of at most $1{,}024$ tokens
each; LLM-as-a-Coach responses are limited to $8{,}192$ tokens throughout. We
optimize all LLM-as-a-Coach models with AdamW~\citep{adam}, using a constant
learning rate of $10^{-6}$.

We train three variants of the LLM-as-a-Coach, each initialized from the same
checkpoint as its frozen actor.
\textbf{(1) Same-instance experiential learning ($K{=}2$)} maximizes the
same-instance reward (\Eqref{eq:same-reward}) and is trained for $100$ steps on
math (Qwen3-1.7B/4B/8B), FrozenLake (Qwen3-1.7B), and Sokoban (Qwen3-4B).
\textbf{(2) Same-instance experiential learning ($K{=}10$)} maximizes the same
reward across ten actor responses: one initial response followed by nine guided
responses. At each iteration, one of the eight trajectories
$(e^{(k)}, y^{(k+1)})$ is drawn uniformly to seed the next iteration. Training
runs for $100$ steps on math (Qwen3-1.7B/4B/8B) and Sokoban
(Qwen3-4B-Instruct-2507), and $200$ steps on FrozenLake (Qwen3-1.7B).
\textbf{(3) Cross-instance experiential learning ($K{=}2$)} maximizes the
cross-instance reward (\Eqref{eq:meta-reward}), where the probe set
$\mathcal{P}$ holds eight instances for math and seven for text-games. Training
runs for $100$ steps on math and FrozenLake and $200$ steps on Sokoban. These
budgets bring the reward curves close to convergence. Full configurations are
provided in Appendix~\ref{app:exp_detail_train}.

\paragraph{Baselines}
We compare L2C against three baselines that share the same frozen actor and
evaluation protocol.
\textbf{(1) Base Model.} The actor responds in a single pass, without
experiential knowledge.
\textbf{(2) Self-Refinement.} The actor revises its answer in a second pass
conditioned on its full initial trajectory, with no separate LLM-as-a-Coach and
no experiential knowledge~\citep{selfrefine,rl4f}.
\textbf{(3) LLM-as-a-Coach (w/o training).} The same experiential learning
procedure as L2C, but with the LLM-as-a-Coach left untrained; we also refer to
this baseline as the untrained coach.

\paragraph{Evaluation}
Each evaluation rollout consists of an initial response, experiential knowledge
extraction, and a guided response. We report two accuracies, mirroring the two
rewards of \Cref{sec:method}.

The \emph{same-instance accuracy} scores the guided response on the instance the
experiential knowledge was extracted from (\Eqref{eq:same-reward}). At
iteration $k$,
\begin{equation}
\mathrm{Acc}^{\text{same}}_k \;=\; \mathbb{P}_{(x, y^{(1)}) \sim \mathcal{D}}\!\left[\, y^{(k)} \text{ is correct} \,\right] ,
\label{eq:acc}
\end{equation}
the fraction of instances the actor answers correctly at its $k$-th response.

The \emph{cross-instance accuracy} instead applies knowledge extracted from a
source instance $x$ to the disjoint probe set $\mathcal{P}$
(\Eqref{eq:meta-reward}):
\begin{equation}
\mathrm{Acc}^{\text{cross}} \;=\; \mathbb{E}_{(x, y) \sim \mathcal{D},\; z \sim \mathcal{P}}\!\left[\, \mathcal{V}(z, y'_{z}) \,\right] ,
\qquad y'_{z} \sim \pi_{\mathrm{actor}}(\cdot \mid z,\, e),
\label{eq:acc-cross}
\end{equation}
measuring how well the extracted knowledge transfers to unseen instances.
\begin{table}[!htbp]
\centering
\small
\begin{tabular}{@{}lllc@{}}
\toprule
\textbf{Model} & \textbf{Task} & \textbf{Method} & \makecell{\textbf{Same-Instance}\\\textbf{Accuracy}} \\
\midrule
\multirow{4}{*}{Qwen3-1.7B} & \multirow{4}{*}{FrozenLake}
 & Base Model                          & 7.4 \\
 & & Self-Refinement~\citep{selfrefine,rl4f}  & 36.5 \\[4pt]
 & & LLM-as-a-Coach (w/o training) & 23.4 \\
 & & \textbf{LLM-as-a-Coach + L2C} & \textbf{65.4} \\
\midrule
\multirow{4}{*}{Qwen3-4B} & \multirow{4}{*}{Sokoban}
 & Base Model                          & 3.8 \\
 & & Self-Refinement                   & 10.6 \\[4pt]
 & & LLM-as-a-Coach (w/o training) & 7.6 \\
 & & \textbf{LLM-as-a-Coach + L2C} & \textbf{23.6} \\
\specialrule{\lightrulewidth}{\aboverulesep}{1.5pt}
\specialrule{\lightrulewidth}{0pt}{\belowrulesep}
\multirow{4}{*}{Qwen3-1.7B} & \multirow{4}{*}{Math}
 & Base Model                          & 61.4 \\
 & & Self-Refinement                   & 54.3 \\[4pt]
 & & LLM-as-a-Coach (w/o training) & 66.1 \\
 & & \textbf{LLM-as-a-Coach + L2C} & \textbf{67.8} \\
\midrule
\multirow{4}{*}{Qwen3-4B} & \multirow{4}{*}{Math}
 & Base Model                          & 67.7 \\
 & & Self-Refinement                   & 71.4 \\[4pt]
 & & LLM-as-a-Coach (w/o training) & 74.4 \\
 & & \textbf{LLM-as-a-Coach + L2C} & \textbf{78.6} \\
\midrule
\multirow{4}{*}{Qwen3-8B} & \multirow{4}{*}{Math}
 & Base Model                          & 75.8 \\
 & & Self-Refinement                   & 77.8 \\[4pt]
 & & LLM-as-a-Coach (w/o training) & 81.9 \\
 & & \textbf{LLM-as-a-Coach + L2C} & \textbf{82.9} \\
\bottomrule
\end{tabular}
\vspace{0.2cm}
\caption{Results of L2C and baselines under the same-instance reward
(\Eqref{eq:same-reward}) on text-games and math. L2C consistently enhances base
models' ability to learn from prior interactions, improving guided
accuracy across tasks and scales.}
\label{tab:vanilla}
\end{table}
\subsection{Results}
\label{sec:vanilla-math}

We present L2C results on math and text-games in Table~\ref{tab:vanilla},
reporting the same-instance accuracy $\mathrm{Acc}^{\text{same}}_2$ with the LLM-as-a-Coach trained
using experiential learning at $K{=}2$ under the same-instance
reward (\Eqref{eq:same-reward}). In
all settings the actor and LLM-as-a-Coach are initialized from the same Qwen3 model; the
actor is frozen throughout, with only the LLM-as-a-Coach updated.

L2C delivers its largest gains on text-games. Unlike math, where the problem
statement fully specifies the task, text-games have latent rules that must be
inferred from interaction. Through training, the LLM-as-a-Coach learns to infer these
rules from the actor's trajectory and distills them into higher-quality
experiential knowledge; conditioned on it, the actor acts more effectively and
achieves better performance.

On math, L2C consistently improves over the untrained coach across all model
sizes. The LLM-as-a-Coach distills the actor's own reasoning trajectory into experiential
knowledge that concentrates on its most useful steps, steering the actor toward
correct solutions. L2C also outperforms
Self-Refinement, underscoring the value
of \emph{coaching}: rather than re-feeding the actor its own full trajectory, the
LLM-as-a-Coach distills it into a compact, actionable experiential knowledge snippet
that more effectively guides the actor's response.

\subsection{LLM-as-a-Coach Yields Transferable Experiential Knowledge}
\label{sec:meta-transfer}

\begin{figure}[!htbp]
\centering
\includegraphics[width=\linewidth]{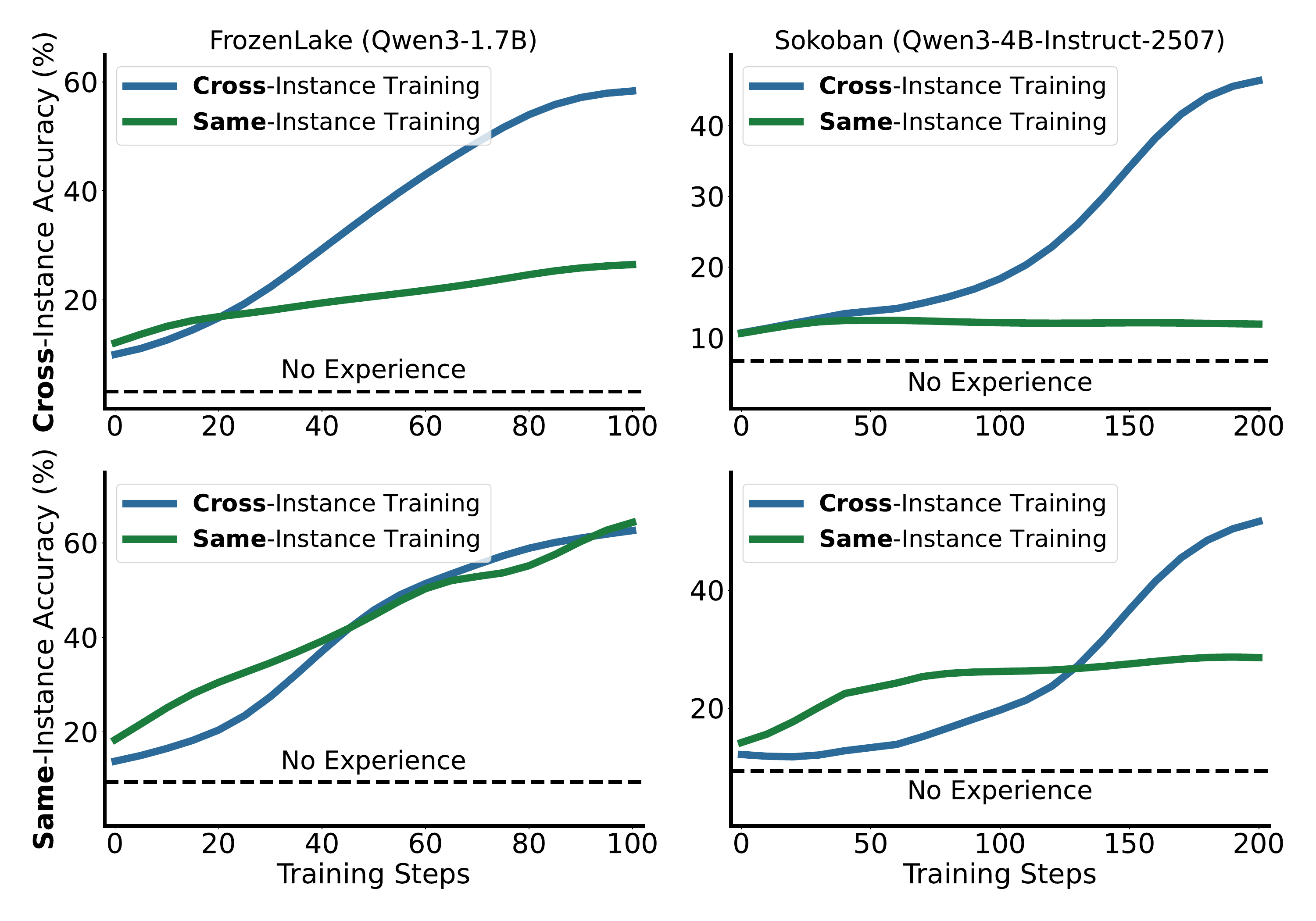}
\caption{Same- and cross-instance accuracy of $K{=}2$ LLM-as-a-Coach trained with
the same-instance reward or the cross-instance reward. Cross-instance reward
training produces substantially more transferable experiential knowledge on
both FrozenLake (left) and Sokoban (right).}
\label{fig:meta-transfer}
\end{figure}

We train $K{=}2$ LLM-as-a-Coach with either the same-instance reward
(\Eqref{eq:same-reward}) or the cross-instance reward
(\Eqref{eq:meta-reward}), and evaluate both using same-instance accuracy and cross-instance
accuracy. We consider FrozenLake with Qwen3-1.7B and Sokoban with
Qwen3-4B-Instruct-2507. The bare actor without experiential knowledge serves as
the baseline. Appendix~\ref{app:meta-eval} provides full evaluation details.

\Cref{fig:meta-transfer} shows that training with the cross-instance reward
yields large, transferable gains: cross-instance accuracy climbs far
above the no-experience baseline, and the improvement generalizes across
environments and model sizes: FrozenLake with Qwen3-1.7B and Sokoban with
Qwen3-4B-Instruct-2507 alike. The cross-instance reward also transfers better than
the same-instance reward: on FrozenLake the cross-instance-trained LLM-as-a-Coach reaches
far higher accuracy than the same-instance-trained one, and a similar gap
emerges on Sokoban. Rewarding knowledge for helping other instances thus
pushes the LLM-as-a-Coach to extract environment-general rules rather than memorize
instance-specific details, positioning L2C as a mechanism for distilling reusable,
environment-general knowledge that generalizes to unseen instances. On math, we observe that fully specified problem statement
leaves little latent structure to share across instances. We report these
results in Appendix~\ref{app:math-meta}.

\subsection{Scaling Up Experiential Learning Iterations}
\label{sec:kiter}

\begin{figure}[!htbp]
\centering
\includegraphics[width=0.96\linewidth]{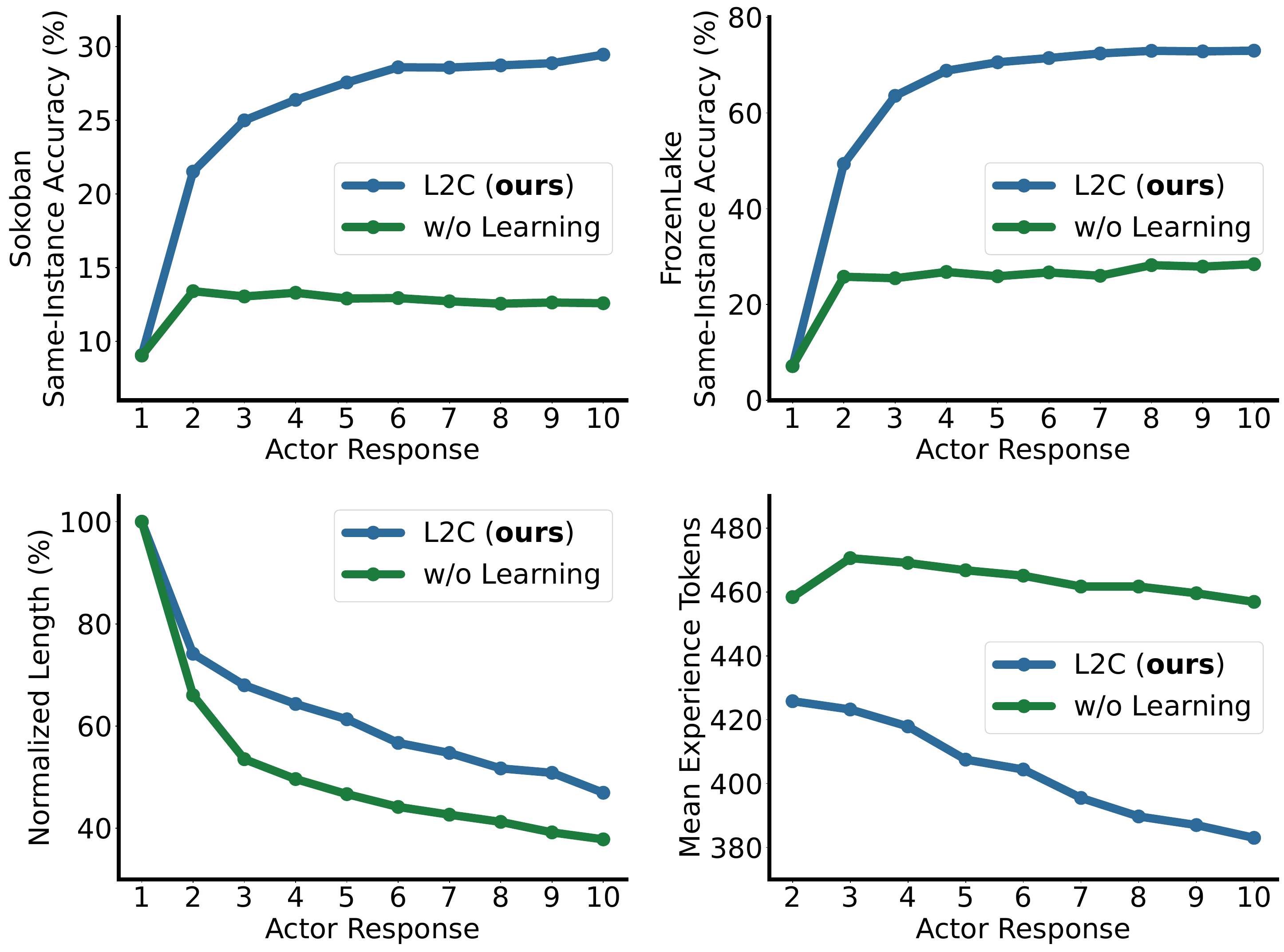}
\caption{Experiential learning with $K{=}10$ under the same-instance reward.
\textbf{Top}: same-instance accuracy on Sokoban (left) and FrozenLake (right).
\textbf{Bottom}: normalized actor-response length and experiential-knowledge length on Sokoban. The trained LLM-as-a-Coach improves
accuracy across iterations while producing progressively shorter experiential
knowledge and actor responses.}
\label{fig:kiter-4b}
\vspace{-0.1cm}
\end{figure}

\Cref{sec:vanilla-math} reports the $K{=}2$ case of experiential learning; here
we scale $K$ to $10$ under the same-instance reward (\Eqref{eq:same-reward}). We
train the LLM-as-a-Coach under the $K{=}10$ objective
(Algorithm~\ref{alg:l2c}) and compare it against an untrained
LLM-as-a-Coach, reporting $\mathrm{Acc}^{\text{same}}_k$ at
every actor response $k \in \{1, \dots, 10\}$; the two settings coincide at
$k{=}1$, the initial unguided response. \Cref{fig:kiter-4b} plots per-response accuracy for
Qwen3-4B-Instruct-2507 on Sokoban and Qwen3-1.7B on FrozenLake, together with
the actor response length and experiential knowledge length on Sokoban.
Appendix~\ref{app:kiter-table} reports $\mathrm{Acc}^{\text{same}}_{10}$ across
all three model sizes and both tasks.

Additional experiential learning iterations scale accuracy monotonically, and training the
LLM-as-a-Coach amplifies the effect. In \Cref{fig:kiter-4b} (top), the trained
curve rises steadily across all ten actor responses, whereas the untrained coach
plateaus after the first guided response. The gain holds across all three model
sizes and both tasks (Appendix~\ref{app:kiter-table}), where L2C improves
$\mathrm{Acc}^{\text{same}}_{10}$ over both the bare actor and the untrained
coach. Iteration is therefore the strongest axis of test-time scaling that L2C
unlocks.

Accuracy rises even as the actor's responses grow shorter.
\Cref{fig:kiter-4b} (bottom-left) shows that the actor's responses contract
monotonically with $k$ under both coaches, so later responses cost less than
the first. Progressively refined experiential knowledge
thus lets the actor reach higher accuracy at a lower inference cost.

The trained LLM-as-a-Coach also produces shorter experiential knowledge.
\Cref{fig:kiter-4b} (bottom-right) shows that its output sits below the
untrained baseline at every response and keeps contracting with $k$, whereas
the untrained curve stays nearly flat. Because the LLM-as-a-Coach consumes the
prior knowledge together with the actor's fresh response, this
contraction indicates that each extraction step \emph{refines} the previous knowledge
rather than restarting from scratch, making it progressively more direct and
precise.

\begin{wraptable}{r}{0.51\textwidth}
\centering
\vspace{-0.4cm}
\small
\setlength{\tabcolsep}{5pt}
\begin{tabular}{@{}llc@{}}
\toprule
\multirow[c]{2}{*}{\raisebox{-0.8ex}{\textbf{Model}}}
& \multirow[c]{2}{*}{\raisebox{-2ex}{%
    \makecell[l]{\textbf{Compute}\\\textbf{Allocation}}}}
& \textbf{DAPO} \\
\cmidrule(lr){3-3}
& & \makecell{\textbf{Accuracy (\%)}} \\
\midrule
\multirow{3}{*}{Qwen3-1.7B}
& Decoding 32k     & 67.9 \\
& L2C, 2 iterations   & 67.8 \\
& \textbf{L2C, 10 iterations} & \textbf{71.0} \\
\midrule
\multirow{3}{*}{Qwen3-4B}
& Decoding 32k     & 77.0 \\
& L2C, 2 iterations   & 78.7 \\
& \textbf{L2C, 10 iterations} & \textbf{82.6} \\
\bottomrule
\end{tabular}
\caption{Compute scaling on DAPO with a frozen actor. ``L2C, $K$ iterations''
denotes running a trained coach for $K$ iterations.
Experiential learning uses additional inference compute more effectively than doubling the decoding budget.}
\label{tab:compute-scaling}
\vspace{-0.6cm}
\end{wraptable}

Experiential learning uses additional inference compute more effectively than simply enlarging
the actor's decoding budget. Compared with a doubled $32$k decoding budget on
math (\Cref{tab:compute-scaling}), a single guided response at $K{=}2$ already
matches the extended-budget result for Qwen3-1.7B and surpasses it for
Qwen3-4B; increasing to $K{=}10$ yields further gains across all evaluated
model sizes. Iterative
coaching thus converts inference compute into repeated opportunities for
course correction---a form of test-time scaling that decoding-budget expansion
alone does not provide. These results suggest that \emph{where}
the extra compute is spent matters as much as \emph{how much} is spent.

\begin{figure}[H]
\centering
\includegraphics[width=0.96\linewidth]{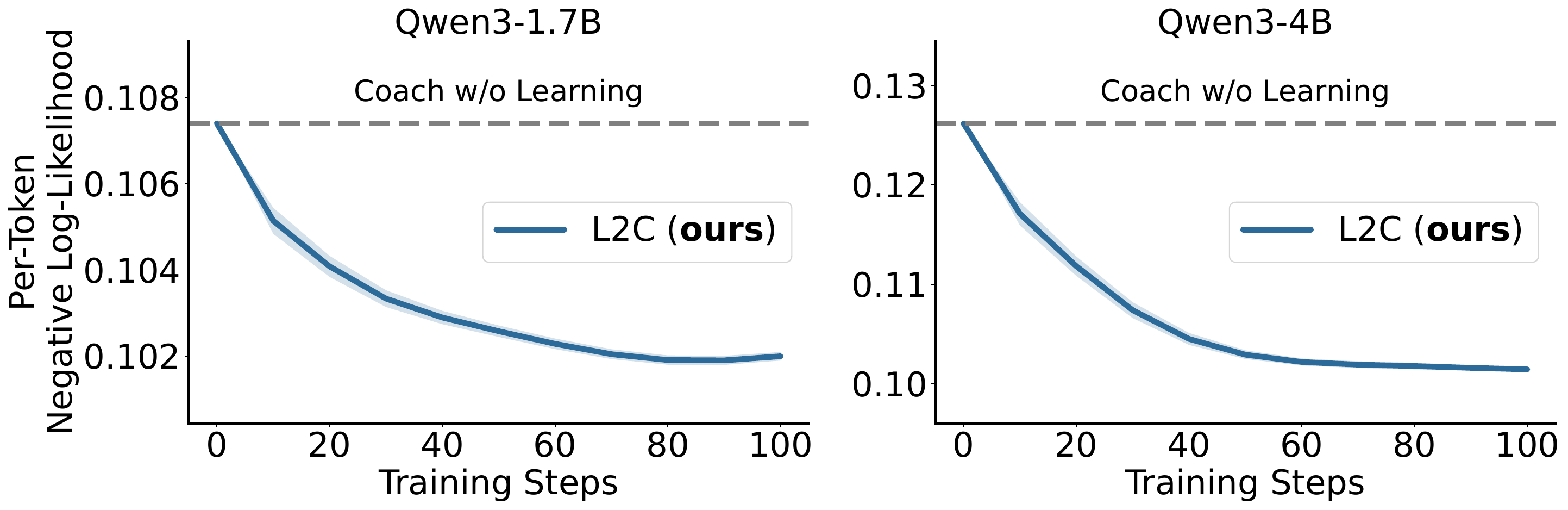}
\caption{Per-token negative log-likelihood of the frozen actor's guided responses across
LLM-as-a-Coach training steps on the DAPO test split, for Qwen3-1.7B (left)
and Qwen3-4B (right). L2C training consistently reduces the negative log-likelihood below the untrained
LLM-as-a-Coach baseline, indicating that the trained LLM-as-a-Coach's guidance is better adapted to the actor's own policy.}
\label{fig:confidence}
\end{figure}

\subsection{LLM-as-a-Coach Adapts Its Guidance to the Actor}
\label{sec:confidence}

Beyond accuracy, we ask whether the LLM-as-a-Coach tailors the experiential
knowledge it extracts to the specific actor it coaches. We measure the per-token
negative log-likelihood that the frozen actor assigns to its own guided
response $y'$ when conditioned on the experiential knowledge $e$,
\begin{equation}
\text{Negative Log-Likelihood} \;=\; -\frac{1}{|y'|} \sum_{t} \log \pi_{\mathrm{actor}}\!\left(y'_t \mid y'_{<t},\, x,\, e\right),
\qquad y' \sim \pi_{\mathrm{actor}}(\cdot \mid x, e),
\label{eq:nll}
\end{equation}
where $y'$ is drawn from the actor's own on-policy distribution. A lower negative
log-likelihood means the guided response lies in a higher-probability region of
the actor's own output distribution, indicating that $e$ is well matched to this
particular actor.

\Cref{fig:confidence} plots the negative log-likelihood on the DAPO test split
across LLM-as-a-Coach training steps for Qwen3-1.7B and Qwen3-4B, with the
untrained LLM-as-a-Coach (step $0$) as the reference. L2C training consistently
drives it below this baseline, and the gap widens over training.

We attribute this to two coupled effects. First, the trained LLM-as-a-Coach
supplies clearer intermediate cues, so the actor commits more decisively to its
guided response. Second, because the LLM-as-a-Coach is optimized against the
actor's own on-policy rollouts, its experiential knowledge comes to elicit
reasoning that stays close to the actor's familiar language and reasoning
templates rather than pushing it off-distribution. The LLM-as-a-Coach therefore
adapts its guidance to the specific actor rather than producing generic advice.
This is consistent with prior work relating response likelihood to the quality
and compatibility of the provided context~\citep{openclawrl}.

\subsection{LLM-as-a-Coach Generalizes Beyond Verifiable Tasks}
\label{sec:generalization}

We train the LLM-as-a-Coach on DAPO math with the same-instance reward and
evaluate its transfer to AIME~2026, text-games, and IFEval, with DAPO as the
in-distribution reference. All four benchmarks follow the same $K{=}2$
protocol as the main results---an initial response, experiential knowledge
extraction, and a guided response---and are scored on the guided response:
DAPO, AIME~2026, and the text-games by same-instance accuracy, and IFEval by
strict instruction-following accuracy. Full evaluation protocols are provided in
Appendices~\ref{app:ood-eval} and~\ref{app:ifeval_ood}.
\begin{figure}[!htbp]
\centering
\includegraphics[width=0.9\linewidth]{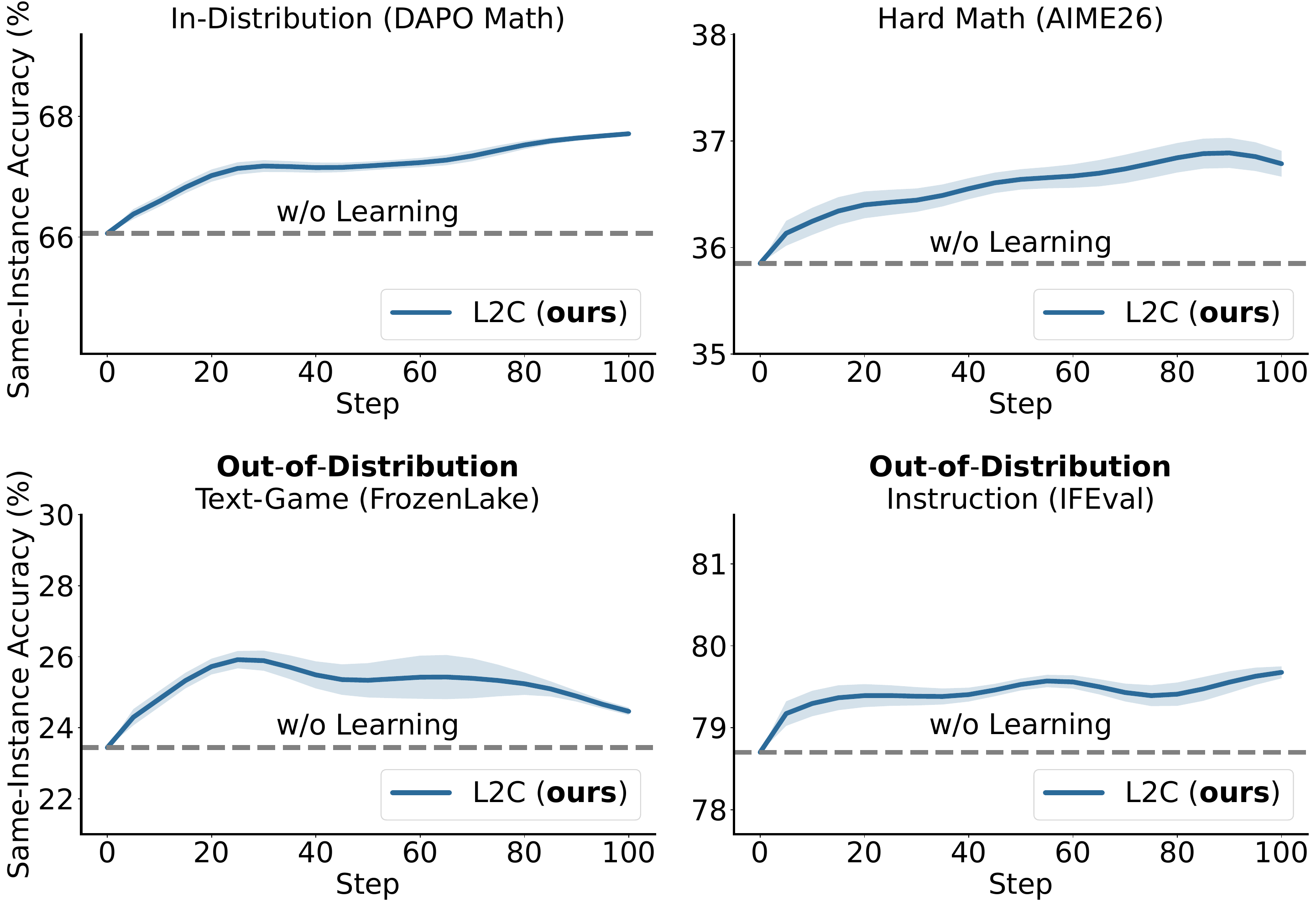}
\caption{Performance across L2C training steps for LLM-as-a-Coach trained on DAPO
math. Results use Qwen3-1.7B on DAPO (top-left), AIME~2026 (top-right), and
FrozenLake (bottom-left), and Qwen3-4B on IFEval (bottom-right). Coaching
ability trained with verifiable math rewards transfers to out-of-distribution reasoning, interaction,
and instruction-following tasks.}
\label{fig:id-ood-1.7b}
\end{figure}
L2C improves over the untrained LLM-as-a-Coach on all four benchmarks
(\Cref{fig:id-ood-1.7b}): harder math, interactive
text-games, and instruction following, in addition to the in-distribution
reference. \Cref{tab:ood-step100} confirms positive transfer across model sizes. Training on verifiable math
rewards therefore develops a general ability to extract and communicate
actionable knowledge from prior trajectories, rather than a skill tied to the
training domain.

\section{Related Work}

\paragraph{Learning from Experience}
Language agents can improve by reusing information acquired through prior
interactions~\citep{silversutton,earlyexperience}. Existing approaches prompt
agents to reflect on failures~\citep{reflexion,selfrefine}, retain knowledge in
external memory~\citep{expel,generativeagents,coala}, construct reusable skill
or workflow libraries~\citep{voyager,agentworkflowmemory}, or optimize prompts
and programs through textual feedback~\citep{textgrad,gepa,feedback-descent}.
These methods demonstrate the value of experiential knowledge, but typically
rely on fixed prompting strategies, hand-designed memory operations, or
heuristic extraction procedures. L2C instead treats experience extraction as
a trainable decision problem: an LLM-as-a-Coach is optimized using the downstream reward
of the frozen actor's guided response. Its same-instance and cross-instance
rewards further distinguish knowledge that helps revisit a particular
problem from knowledge that transfers across instances.

\paragraph{Learned Critics and Teachers}
Auxiliary language models have been used to evaluate or improve model outputs,
including AI-feedback critics~\citep{rlaif}, LLM-as-a-Judge
systems~\citep{mtbench}, process reward models~\citep{prm,mathshepherd}, and
models that identify errors in generated code or reasoning~\citep{criticgpt}.
Teacher models can also revise instruction-tuning data~\citep{coachlm} or
provide natural-language feedback for refinement~\citep{rl4f,coachel}. Related
RL approaches optimize an actor using a fixed critic~\citep{critique-grpo,golf},
train a critic for refinement utility~\citep{rco}, or jointly evolve the critic
and actor~\citep{echo}. In contrast, L2C freezes the actor and optimizes only
the LLM-as-a-Coach using rewards from guided responses, learning to distill
prior trajectories into actor-specific experiential knowledge that transfers
across instances, rather than instance-bound critiques.

\FloatBarrier
\section{Conclusion}

In this work, we introduced Learning to Coach (L2C), a framework that trains a
dedicated LLM-as-a-Coach to extract actionable experiential knowledge from a
frozen actor's previous trajectory. By optimizing the LLM-as-a-Coach with the
reward of the actor's guided response, L2C improves downstream performance
without modifying the actor. Across mathematical reasoning and interactive
text-games, L2C consistently outperforms self-refinement and an untrained
LLM-as-a-Coach. We further showed that the cross-instance reward induces
transferable knowledge when instances share latent structure, and that scaling up
experiential learning iterations provides more effective test-time scaling than
enlarging the actor's decoding budget. The trained LLM-as-a-Coach also produces compact
experiential knowledge and transfers to out-of-distribution tasks. Moreover, it
adapts its guidance to the specific actor it coaches. These results establish
LLM-as-a-Coach training as a modular approach for enabling language models to
benefit from prior interactions while preserving the capabilities of the models
being coached.

\bibliographystyle{alpha}
\bibliography{opcd}

\newpage
\appendix

\section{Training Details}
\label{app:train_detail}

\subsection{Datasets}
\label{app:exp_detail_dataset}

Our training data span three task domains: mathematical problem solving and two
interactive, text-only games. For mathematics, we use DAPO-Math-17K~\citep{dapo},
which includes roughly 14K English-language problems whose final numerical
answers can be automatically verified. For interactive reasoning, we adopt the
Frozen Lake and Sokoban environments provided by TextArena~\citep{textarena}.

In Frozen Lake, the agent navigates a $3{\times}3$ board containing two holes
and must find a safe route to the goal. Sokoban is instantiated on a
$6{\times}6$ board with a single box; the agent must push the box onto its
target while avoiding holes, walls, and irreversible deadlocks. Following
prior work~\citep{tencentgame}, we omit some of the game instructions so that
the agent must discover the missing mechanics through interaction. At every
turn, TextArena converts the current board configuration into a textual
observation, and the language model responds with an action, resulting in a
multi-turn trajectory.

\subsection{Training Configuration}
\label{app:exp_detail_train}

We optimize the LLM-as-a-Coach with GRPO~\citep{grpo} while keeping the actor frozen.
For each source trajectory, we sample $n{=}8$ candidate experiential-knowledge
snippets from the LLM-as-a-Coach. Each candidate conditions an independent guided response
by the actor, and the resulting eight rewards form one GRPO
advantage-normalization group. All LLM-as-a-Coach models are optimized with
AdamW~\citep{adam} using a constant learning rate of $10^{-6}$.

\paragraph{Math.}
For math, each mini-batch contains $64$ source problems for the Qwen3-1.7B and
Qwen3-4B LLM-as-a-Coach models and $128$ source problems for the Qwen3-8B LLM-as-a-Coach. Actor
responses, both initial and guided, are limited to $16{,}384$
tokens, and LLM-as-a-Coach responses are limited to $8{,}192$ tokens. Under the
same-instance reward, each candidate conditions an independent guided response on
the source problem, and a deterministic answer verifier assigns a binary
correctness reward.

\paragraph{Text-games.}
For text-games, each mini-batch contains $64$ environment seeds. An initial or
guided response is a playthrough of at most five interaction turns, with actor
responses limited to $1{,}024$ tokens per turn. LLM-as-a-Coach responses are limited to
$8{,}192$ tokens. Under the same-instance reward, each candidate conditions an
independent guided playthrough using the same environment seed as the
initial response, and the binary win/loss outcome provides the reward.

\paragraph{Training variants.}
We train three variants of the LLM-as-a-Coach, each initialized from the same base
checkpoint as its corresponding actor.

\begin{itemize}[leftmargin=1.5em]
    \item \textbf{Experiential learning with $K{=}2$.}
    We use the same-instance reward with $K{=}2$, corresponding to one initial
    response, one extraction step, and one guided response. We train one LLM-as-a-Coach for
    each task--model pair for $100$ gradient steps: math with
    Qwen3-1.7B/4B/8B, FrozenLake with Qwen3-1.7B, and Sokoban with Qwen3-4B.

    \item \textbf{Experiential learning with $K{=}10$.}
    We use the same-instance reward with $K{=}10$, corresponding to one
    initial response followed by nine guided responses. At
    each iteration, the LLM-as-a-Coach samples eight candidate updates to the experiential
    knowledge, and each candidate is evaluated through an independent guided
    response. One of the eight candidates is then selected uniformly, together
    with its corresponding actor trajectory, to provide the state for the next
    iteration. We train the math LLM-as-a-Coach models and the
    Qwen3-4B-Instruct-2507 Sokoban LLM-as-a-Coach for $100$ gradient steps and the
    Qwen3-1.7B FrozenLake LLM-as-a-Coach for $200$ gradient steps.

    \item \textbf{Cross-instance experiential learning.}
    We use $K{=}2$ and replace the same-instance reward with the cross-instance
    reward in \Eqref{eq:meta-reward}. Each step uses $64$ source instances and
    a probe set disjoint from the sources. Each candidate is applied
    independently to every probe instance, and its reward is the mean
    guided accuracy over the probe set. We use eight probes for math, shared
    across all source groups within a step, and seven per-group probes for
    text-games. We train the math and FrozenLake LLM-as-a-Coach models for $100$
    gradient steps and the Qwen3-4B-Instruct-2507 Sokoban LLM-as-a-Coach for $200$
    gradient steps.
\end{itemize}

\Cref{tab:train-steps} summarizes the number of GRPO gradient steps in each
setting.

\begin{table}[!htbp]
\centering
\small
\begin{tabular}{@{}lllc@{}}
\toprule
\textbf{Reward} & \textbf{Protocol} & \textbf{Task (model)}
& \textbf{Steps} \\
\midrule
\multirow{3}{*}{Same-instance}
& \multirow{3}{*}{$K{=}2$ (\Cref{sec:vanilla-math})}
& Math (Qwen3-1.7B / 4B / 8B)       & 100 \\
& & FrozenLake (Qwen3-1.7B)          & 100 \\
& & Sokoban (Qwen3-4B)               & 100 \\
\midrule
\multirow{3}{*}{Same-instance}
& \multirow{3}{*}{$K{=}10$ (\Cref{sec:kiter})}
& Math (Qwen3-1.7B / 4B / 8B)       & 100 \\
& & Sokoban (Qwen3-4B-Instruct-2507) & 100 \\
& & FrozenLake (Qwen3-1.7B)          & 200 \\
\midrule
\multirow{3}{*}{Cross-instance}
& \multirow{3}{*}{$K{=}2$ (\Cref{sec:meta-transfer})}
& Math (Qwen3-1.7B / 4B / 8B)       & 100 \\
& & FrozenLake (Qwen3-1.7B)          & 100 \\
& & Sokoban (Qwen3-4B-Instruct-2507) & 200 \\
\bottomrule
\end{tabular}
\vspace{0.2cm}
\caption{\textbf{L2C training configurations.}
We report the number of GRPO optimizer steps for same-instance experiential
learning ($K{=}2$), same-instance experiential learning ($K{=}10$), and
cross-instance experiential learning ($K{=}2$). All runs use a frozen actor, eight LLM-as-a-Coach
samples per source trajectory, AdamW, and a constant learning rate of $10^{-6}$.}
\label{tab:train-steps}
\end{table}

\subsection{Prompt Templates}
\label{app:exp_detail_templates}
L2C uses two families of prompts, matching the two rewards of
\Cref{sec:method}: the \emph{same-instance} prompts behind the main results
(\Cref{sec:vanilla-math,sec:kiter}), and the \emph{cross-instance} prompts
behind the transfer experiment (\Cref{sec:meta-transfer}). We list both below.

\paragraph{Same-instance prompts.}
For experiential knowledge update on the math dataset, we use the prompt template in \Cref{fig:exp_accum_temp_math}.

\begin{figure}[!htbp]
    \begin{tcolorbox}
    You are solving a problem. Below are your previous attempts and the notes you took while solving. Refine the notes so your next attempt is more likely to be correct. \\ \\
    Problem and your latest attempt: \\
    \{LATEST\_EXPERIENCE\} \\ \\
    Your existing notes on this problem (from earlier attempts): \\
    \# Notes \\
    \{PREVIOUS\_EXPERIENCE\} \\ \\
    Your task: \\
    - Critique the latest attempt. Where did the reasoning go wrong, or what was left unverified? If the answer seems correct, what would make you more sure? \\
    - Update the notes so they capture: (a) what you have already tried and why it did or did not work, (b) the most promising direction for the next attempt, (c) any intermediate results worth keeping (lemmas, simplifications, candidate answers with confidence). \\
    - The notes will REPLACE the previous notes, not be appended. Keep them concise. They will be re-read at every attempt. \\ \\
    After reasoning step by step, output the final notes in exactly this format: \\ \\
    \# Notes \\
    - ... \\
    - ...
    \end{tcolorbox}
    \caption{The prompt template for the LLM-as-a-Coach's notes update on the math dataset.}
    \label{fig:exp_accum_temp_math}
\end{figure}

The LLM-as-a-Coach's output replaces the previous notes verbatim (no per-line item extraction).

For experiential knowledge update on text-based games, we use the prompt template in \Cref{fig:exp_accum_game}.

\begin{figure}[!htbp]
    \begin{tcolorbox}
    You are playing a game. Below is your latest playthrough and the notes you took while playing. Refine the notes so your next attempt is more likely to win. \\ \\
    Interaction history (the game environment (input) and your response and action (output)): \\
    \{LATEST\_EXPERIENCE\} \\ \\
    Your existing notes on this game (from earlier attempts): \\
    \# Notes \\
    \{PREVIOUS\_EXPERIENCE\} \\ \\
    Your task: \\
    - Critique the latest playthrough. Which actions moved you toward or away from the goal, what feedback did the environment give, and where did the attempt fail? \\
    - Update the notes so they capture: (a) confirmed facts about this map (goal location, hazards/walls, what each action does), (b) action sequences that worked or failed and why, (c) the most promising plan for the next attempt. \\
    - The notes will REPLACE the previous notes, not be appended. Keep them concise. They will be re-read at every attempt. \\ \\
    After reasoning step by step, output the final notes in exactly this format: \\ \\
    \# Notes \\
    - ... \\
    - ...
    \end{tcolorbox}
    \caption{The prompt template for the LLM-as-a-Coach's notes update on text-based games.}
    \label{fig:exp_accum_game}
\end{figure}

For guided responses on the same problem (or game), we embed the LLM-as-a-Coach's notes with the prompt template in \Cref{fig:exp_solve}.

\begin{figure}[!htbp]
    \begin{tcolorbox}
    You have been working on the problem below. Here are your notes from previous attempts; use them to avoid past mistakes and find the correct answer. \\ \\
    \# Notes \\
    \{experience\} \\ \\
    Solve the problem. You may continue from a promising direction in the notes or start over if needed; the notes are guidance, not constraints. \\ \\
    Problem: \\
    \{prompt\}
    \end{tcolorbox}
    \caption{The prompt template for a guided response with the LLM-as-a-Coach's notes.}
    \label{fig:exp_solve}
\end{figure}

\paragraph{Cross-instance prompts.}
The cross-instance variant rewards \emph{transferable} experience, so its
prompts extract general, high-level insight that is accumulated into a running
knowledge base and then applied to \emph{new} instances. For experience update
on the math dataset we use \Cref{fig:cross_accum_math}; on text-based games (the
setting of the FrozenLake transfer experiment in \Cref{sec:meta-transfer}) we
use \Cref{fig:cross_accum_game}. The accumulated experience is applied to a new
instance with the solve template in \Cref{fig:cross_solve}.

\begin{figure}[!htbp]
    \begin{tcolorbox}
    You are an AI language model that continuously refines its internal experience. \\ \\
    Here is the latest interaction (the user's question and your answer): \\
    \{LATEST\_EXPERIENCE\} \\ \\
    Here is the previous experience: \\
    \# Experience \\
    \{PREVIOUS\_EXPERIENCE\} \\ \\
    Your task: \\
    Based on the latest interaction and the previous experience, generate an additional experience for future learning. \\ \\
    Rules: \\
    - The experience you generate MUST be formatted strictly as a markdown list where each item starts with ``- EXPERIENCE ITEM:'', one per line: \\
    - EXPERIENCE ITEM: ... \\
    - EXPERIENCE ITEM: ... \\
    - EXPERIENCE ITEM: ... \\
    - The experience you generate will be directly appended to the previous experience. \\
    - The change should introduce a general, high-level, widely applicable insight, not a detail from the specific interaction. The updated experience must remain concise, structured, and meaningful. \\
    - If the new insight conflicts with any previous experience item, you can describe the conflict and provide a resolution in the new item. \\ \\
    After careful reasoning step by step, output the final result in exactly this format: \\ \\
    Additional Experience: \\
    \# Experience \\
    - EXPERIENCE ITEM: ... \\
    - EXPERIENCE ITEM: ... \\
    - EXPERIENCE ITEM: ...
    \end{tcolorbox}
    \caption{The cross-instance prompt template for experience extraction on the math dataset: general, transferable insight accumulated as appended experience items.}
    \label{fig:cross_accum_math}
\end{figure}

\begin{figure}[!htbp]
    \begin{tcolorbox}
    You are an AI language model that continuously refines its internal experience. \\
    Here is the interaction history (the game environment (input) and your response and action (output)): \\
    \{LATEST\_EXPERIENCE\} \\ \\
    Here is the previous experience: \\
    \# Experience \\
    \{PREVIOUS\_EXPERIENCE\} \\ \\
    Your task: \\
    Based on the multi-round interaction history and the previous experience, generate experience for future learning. You should conduct a deep, comparative analysis to infer the game rules and the fundamental principles behind winning and losing. Using the interaction history and environment feedback, hypothesize the game rules and effective winning strategies, and organize these insights into 1--2 concise, high-level, and widely applicable experience items that help the player succeed in the game. \\ \\
    Rules: \\
    - The experience you generate MUST be formatted strictly as a markdown item which starts with ``- EXPERIENCE ITEM:'': \\
    - EXPERIENCE ITEM: ... \\
    - EXPERIENCE ITEM: ... \\
    - The experience you generate will be directly appended to the previous experience. Do not repeat the previous experience. Make sure the newly generated experience is different from the previous experience. \\
    - Your generated experience should be possible rules, instructions or winning strategies for the game. The experience should be generally useful rather than only applicable for the current map (board). \\ \\
    After careful reasoning step by step, output the final result in exactly this format: \\ \\
    Additional Experience (Rules or Strategies): \\
    \# Experience \\
    - EXPERIENCE ITEM: ...
    \end{tcolorbox}
    \caption{The cross-instance prompt template for experience extraction on text-based games: transferable game rules and winning strategies rather than map-specific notes.}
    \label{fig:cross_accum_game}
\end{figure}

\begin{figure}[!htbp]
    \begin{tcolorbox}
    Given previous learned experience: \\
    \# Experience \\
    \{experience\} \\ \\
    Solve the new problem and explain what part of experience you use and how you use it in the reasoning process: \\
    \{prompt\}
    \end{tcolorbox}
    \caption{The cross-instance prompt template for solving a new instance with the accumulated experience.}
    \label{fig:cross_solve}
\end{figure}

\subsection{Examples of Learned Experience}
\label{app:exp-examples}
The two reward variants produce qualitatively different experiential knowledge,
mirroring their prompts. \Cref{fig:exp-same-example} shows a representative
\emph{same-instance} note learned on FrozenLake (Qwen3-1.7B): it is tied to one
specific map, recording exact obstacle coordinates and a memorized step-by-step
solution path. \Cref{fig:exp-cross-example} shows a \emph{cross-instance}
experience from the same setting: it abstracts the interaction into general game
rules and a reusable navigation strategy intended to apply across maps.

\begin{figure}[!htbp]
    \begin{tcolorbox}
    \# Notes \\
    - \textbf{Board Layout}: 3x3 grid with player (P) at (0,0), goal (G) at (2,2), and obstacles (H) at (0,2) and (1,2). Valid moves: up (w), down (s), left (a), right (d). \\
    - \textbf{Optimal Path}: \\
    ~~1. \textbf{Down (s)} from (0,0) to (1,0) $\to$ Establishes vertical path toward the goal. \\
    ~~2. \textbf{Down (s)} from (1,0) to (2,0) $\to$ Progresses toward the goal. \\
    ~~3. \textbf{Right (d)} from (2,0) to (2,1) $\to$ Advances row. \\
    ~~4. \textbf{Right (d)} from (2,1) to (2,2) $\to$ Directly reaches the goal (G). \\
    - \textbf{Critical Failures}: \\
    ~~- Moving \textbf{up (w)} from (0,0) is invalid (out of bounds). \\
    ~~- Moving \textbf{left (a)} from (0,0) is invalid (out of bounds).
    \end{tcolorbox}
    \caption{A \emph{same-instance} experience (v5) learned on FrozenLake (Qwen3-1.7B). The note memorizes one map's layout and a concrete solution path.}
    \label{fig:exp-same-example}
\end{figure}

\begin{figure}[!htbp]
    \begin{tcolorbox}
    \# Experience \\
    - EXPERIENCE ITEM: \textbf{Game Rules and Winning Strategy} \\
    The game involves navigating a 3x3 grid where the player (P) starts at the top-left corner (row 0, column 0) and must reach the goal (G) at the bottom-right corner (row 2, column 2). The grid contains obstacles (H) that block movement. The player can move up, down, left, or right, and must avoid obstacles while reaching the goal. \\ \\
    - EXPERIENCE ITEM: \textbf{Effective Winning Strategy} \\
    The optimal strategy is to move \textbf{down} first to progress toward the goal. From the starting position (0,0), move \textbf{down} to (1,0), then \textbf{right} to (1,1), (1,2), and finally \textbf{down} to (2,2). This path avoids obstacles and ensures the shortest route to the goal. Always prioritize moving \textbf{down} to progress toward the goal, and use \textbf{right} to navigate through empty spaces.
    \end{tcolorbox}
    \caption{A \emph{cross-instance} experience (v4) learned on FrozenLake (Qwen3-1.7B). The experience abstracts the interaction into game rules and a reusable strategy rather than a single map's solution.}
    \label{fig:exp-cross-example}
\end{figure}

\section{Evaluation Details}
\label{app:eval_detail}

\subsection{Evaluation Configuration}
\label{app:eval-config}
For math, we evaluate on the DAPO test split ($1{,}000$ problems, $n{=}16$
rollouts per problem). For text-games, we use $500$ held-out environment seeds
($n{=}8$ playthroughs per seed).

\subsection{Out-of-Distribution Generalization Evaluation}
\label{app:ood-eval}
For the out-of-distribution generalization results of \Cref{sec:generalization}
(\Cref{tab:ood-step100}), we evaluate on AIME~2026 with $30$ problems and
$n{=}128$ rollouts per problem, and on the text-games with $500$ held-out
environment seeds and $n{=}8$ playthroughs per seed. The in-distribution DAPO
reference follows the main evaluation protocol of Appendix~\ref{app:eval-config}.

\begin{table}[t]
\centering
\small
\begin{tabular}{@{}llccc@{}}
\toprule
\makecell{\textbf{Out-of-Distribution}\\\textbf{Task}} & \textbf{Model} & \tabincell{c}{\textbf{$\mathrm{Acc}^{\text{same}}_2$} \\ \textbf{(w/o training)}} & \tabincell{c}{\textbf{$\mathrm{Acc}^{\text{same}}_2$} \\ \textbf{(L2C)}} & \tabincell{c}{\textbf{$\Delta$} \\ \textbf{(pp)}} \\
\midrule
\multirow{2}{*}{AIME26}
 & Qwen3-1.7B                & 35.9 & \textbf{37.4} & $+1.5$ \\
 & Qwen3-8B                  & 61.6 & \textbf{63.4} & $+1.8$ \\
\midrule
\multirow{2}{*}{Text-game}
 & Qwen3-1.7B (FrozenLake)   & 23.4 & \textbf{26.3} & $+2.9$ \\
 & Qwen3-4B  (Sokoban)       & 7.6  & \textbf{9.4}  & $+1.8$ \\
\bottomrule
\end{tabular}
\vspace{0.2cm}
\caption{\textbf{Out-of-distribution transfer of LLM-as-a-Coach models trained on DAPO math.}
Same-instance accuracy (\%) after one guided response on AIME~2026 and
text-game tasks. The actor remains frozen, and $\Delta$ denotes the
percentage-point improvement of L2C over the untrained LLM-as-a-Coach.}
\label{tab:ood-step100}
\end{table}

\subsection{IFEval Out-of-Distribution Evaluation}
\label{app:ifeval_ood}
For the out-of-distribution instruction-following evaluation
(\Cref{sec:generalization}), we evaluate the trained LLM-as-a-Coach--actor system
on IFEval~\citep{ifeval} under the same $K{=}2$ protocol used for math and
text-games: the frozen actor first answers the prompt on its own, the
LLM-as-a-Coach extracts experiential knowledge from the resulting trajectory,
and the actor then produces a guided response conditioned on that knowledge.
Only the guided response is scored. The untrained-coach baseline is step $0$ of
the identical pipeline.

We evaluate on the first $250$ prompts of the IFEval test split as ordered by
\texttt{lm-eval-harness}. For each prompt, the LLM-as-a-Coach samples $n{=}16$
independent experiential-knowledge snippets from the actor's initial response,
using temperature $0.6$, top-$p{=}0.95$, top-$k{=}20$, and an $8{,}192$-token
budget, matching the training-time rollout configuration. Each snippet
conditions one guided response, yielding $250\times16=4{,}000$ guided responses
per checkpoint. In both actor passes the model's chat template is applied as a
single user turn with thinking mode disabled, and decoding is greedy with a
$1{,}280$-token budget. Because IFEval grades literal surface compliance, we
inject the experiential knowledge with a shortened wrapper rather than the
longer training-time template of \Cref{fig:exp_solve}, whose instruction-heavy
preamble by itself depresses prompt-level strict accuracy for the trained and
untrained coach alike. Each guided response is scored by IFEval's prompt-level
strict criterion (\texttt{prompt\_level\_strict\_acc}); we report the per-prompt
mean over the $16$ snippets and then average over the $250$ prompts. All other
hyperparameters follow \texttt{lm-eval-harness} defaults.

\subsection{Cross-Instance Transfer Evaluation}
\label{app:meta-eval}
We evaluate cross-instance transfer using two disjoint pools drawn from the
task: a source pool $\mathcal{S}$ of $|\mathcal{S}|{=}64$ instances and a
held-out probe set $\mathcal{P}$ of $|\mathcal{P}|{=}250$ instances
($16{,}000$ source--probe pairs), used for both FrozenLake (Qwen3-1.7B; the
main-text experiment of \Cref{sec:meta-transfer}, where instances are
environment seeds) and DAPO math (Appendix~\ref{app:math-meta}, where instances are
problems). For each source instance
$s \in \mathcal{S}$, the actor first responds to $s$ without experiential knowledge and the
LLM-as-a-Coach compacts the resulting trajectory into experiential knowledge $e_s$.
At each checkpoint we then report three quantities:
\begin{itemize}[leftmargin=1.5em,nosep,topsep=2pt]
    \item \textbf{Cross-instance accuracy}: apply each $e_s$ to every
    probe instance $z \in \mathcal{P}$ and average the actor's guided accuracy
    over all $|\mathcal{S}|\times|\mathcal{P}|$ source--probe pairs, matching the
    cross-instance reward of \Eqref{eq:meta-reward}.
    \item \textbf{Same-instance guided accuracy}: apply each $e_s$ back to its
    own source instance $s$, mirroring the same-instance setting of the main
    results; it serves as a memorization discriminator.
    \item \textbf{No-experience baseline}: the actor's accuracy on the probe
    instances without any experiential knowledge.
\end{itemize}
Both pools are held fixed across checkpoints, and $\mathcal{P}$ is disjoint from
$\mathcal{S}$ so that transfer is always measured on instances the experiential
knowledge was not extracted from.

\section{Cross-Instance Transfer on Math}
\label{app:math-meta}

\begin{figure}[!htbp]
\centering
\includegraphics[width=\linewidth]{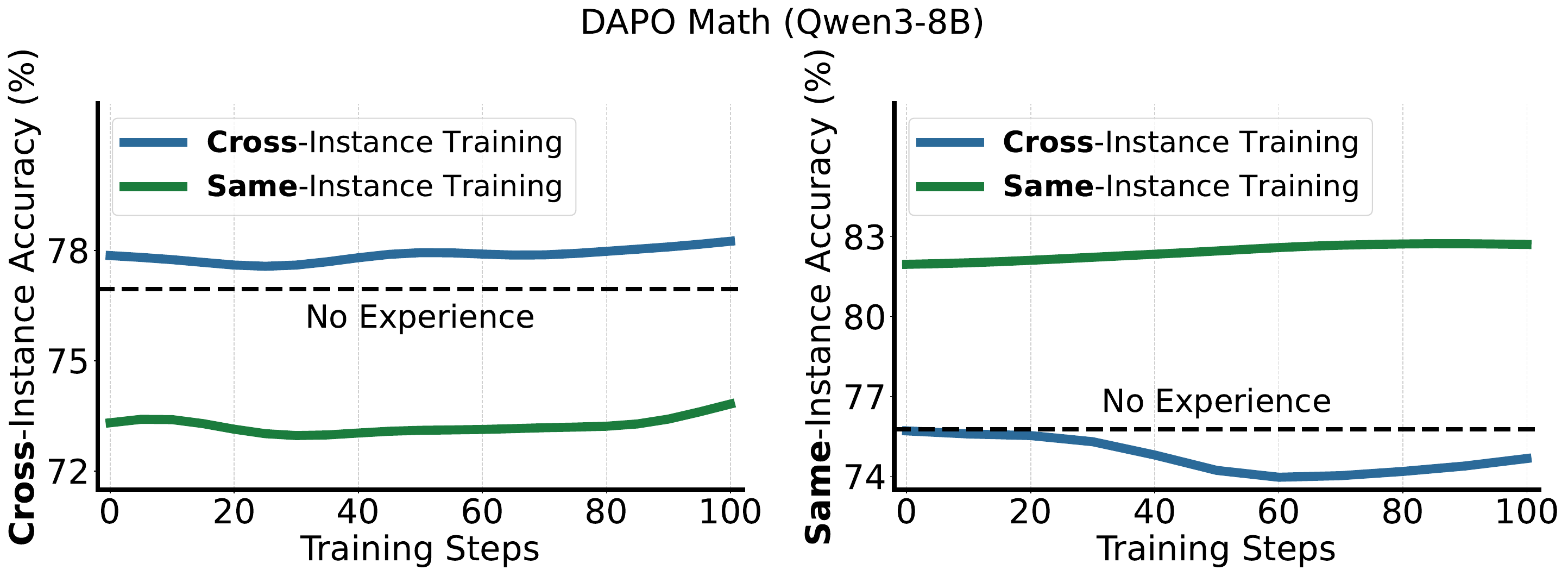}
\caption{LLM-as-a-Coach models trained with the cross-instance or the same-instance reward
on DAPO math (Qwen3-8B), evaluated by cross-instance accuracy (left) and
same-instance accuracy (right). Each reward improves only the accuracy it was
optimized for.}
\label{fig:math-meta-8b}
\end{figure}

Whether experiential knowledge transfers across instances depends on the task:
it is useful only when solving one instance reveals structure that also applies
to others. We probe this by running the identical cross-instance protocol
(Appendix~\ref{app:meta-eval}) on DAPO math for all three model sizes, complementing the
FrozenLake experiment of \Cref{sec:meta-transfer}. \Cref{tab:math-meta} reports
the cross-instance accuracy of the LLM-as-a-Coach before and after L2C training
against the no-experience baseline.

The two tasks sit at opposite ends of this spectrum. On math the extracted
experience transfers little: the cross-instance accuracy stays at or
below the no-experience baseline, and L2C training does not lift it meaningfully
above. On the text-games, by contrast, the same protocol drives cross-instance
accuracy far above the baseline. The gap reflects task structure. A fully-specified math
problem statement leaves little latent, cross-problem structure to transfer, so
experience distilled from one problem's solution rarely helps a different
problem. A text-game's latent rules (hazards, dynamics, and valid moves) are
instead shared across maps and therefore transfer readily. Transferable
experiential knowledge is thus task-dependent: abundant in interactive
environments but scarce in self-contained math problems.
\begin{table}[t]
    \centering
    \small
    \setlength{\tabcolsep}{6pt}
    \begin{tabular}{@{}lccc@{}}
    \toprule
    \multirow{2}{*}{\textbf{Model}}
    & \multirow{2}{*}{\textbf{No Exp.}}
    & \multicolumn{2}{c}{\textbf{Cross-instance transfer}} \\
    \cmidrule(l){3-4}
    & & LLM-as-a-Coach (w/o training) & L2C \\
    \midrule
    Qwen3-1.7B & 64.4 & 49.9 & 51.3 \\
    Qwen3-4B   & 71.0 & 62.5 & 63.7 \\
    Qwen3-8B   & 77.2 & 78.0 & 78.4 \\
    \bottomrule
    \end{tabular}
    \vspace{0.2cm}
    \caption{\textbf{Cross-instance transfer on DAPO math.}
    Cross-instance accuracy (\%) obtained by applying knowledge extracted from
    64 source problems to 250 disjoint probe problems. Cross-problem knowledge provides
    limited benefit, particularly for the smaller actors.}
    \label{tab:math-meta}
\end{table}

\section{Experiential Learning Results at $K{=}10$}
\label{app:kiter-table}
\Cref{tab:kiter-train} reports the final-iteration accuracy $\mathrm{Acc}^{\text{same}}_{10}$
under the $K{=}10$ protocol for the bare actor, the untrained
coach, and L2C, across all three model sizes and both tasks. L2C improves
over both baselines on every combination of model size and task.

\begin{table}[!htbp]
\centering
\small
\begin{tabular}{@{}lllc@{}}
\toprule
\textbf{Model} & \textbf{Task} & \textbf{Method} & \textbf{Accuracy} \\
\midrule
\multirow{3}{*}{Qwen3-1.7B} & \multirow{3}{*}{FrozenLake}
 & Base Model                                 & 7.4 \\
 & & LLM-as-a-Coach w/o training, $\mathrm{Acc}^{\text{same}}_{10}$     & 28.4 \\
 & & \textbf{L2C}, $\mathrm{Acc}^{\text{same}}_{10}$       & \textbf{73.0} \\
\midrule
\multirow{3}{*}{Qwen3-4B-Ins} & \multirow{3}{*}{Sokoban}
 & Base Model                                 & 9.1 \\
 & & LLM-as-a-Coach w/o training, $\mathrm{Acc}^{\text{same}}_{10}$     & 12.6 \\
 & & \textbf{L2C}, $\mathrm{Acc}^{\text{same}}_{10}$       & \textbf{29.5} \\
\specialrule{\lightrulewidth}{\aboverulesep}{1.5pt}
\specialrule{\lightrulewidth}{0pt}{\belowrulesep}
\multirow{3}{*}{Qwen3-1.7B} & \multirow{3}{*}{Math}
 & Base Model                                 & 61.4 \\
 & & LLM-as-a-Coach w/o training, $\mathrm{Acc}^{\text{same}}_{10}$     & 69.9 \\
 & & \textbf{L2C}, $\mathrm{Acc}^{\text{same}}_{10}$       & \textbf{71.0} \\
\midrule
\multirow{3}{*}{Qwen3-4B} & \multirow{3}{*}{Math}
 & Base Model                                 & 67.7 \\
 & & LLM-as-a-Coach w/o training, $\mathrm{Acc}^{\text{same}}_{10}$     & 75.5 \\
 & & \textbf{L2C}, $\mathrm{Acc}^{\text{same}}_{10}$       & \textbf{82.6} \\
\midrule
\multirow{3}{*}{Qwen3-8B} & \multirow{3}{*}{Math}
 & Base Model                                 & 75.8 \\
 & & LLM-as-a-Coach w/o training, $\mathrm{Acc}^{\text{same}}_{10}$     & 82.9 \\
 & & \textbf{L2C}, $\mathrm{Acc}^{\text{same}}_{10}$       & \textbf{85.6} \\
\bottomrule
\end{tabular}
\vspace{0.2cm}
\caption{\textbf{Accuracy under same-instance experiential learning at $K{=}10$.}
The base-actor rows report initial-response accuracy
$\mathrm{Acc}^{\text{same}}_1$, whereas the remaining rows report final accuracy
$\mathrm{Acc}^{\text{same}}_{10}$ after nine guided responses.
L2C outperforms the untrained LLM-as-a-Coach in all evaluated settings.}
\label{tab:kiter-train}
\end{table}
\begin{figure}[!htbp]
\centering
\includegraphics[width=\linewidth]{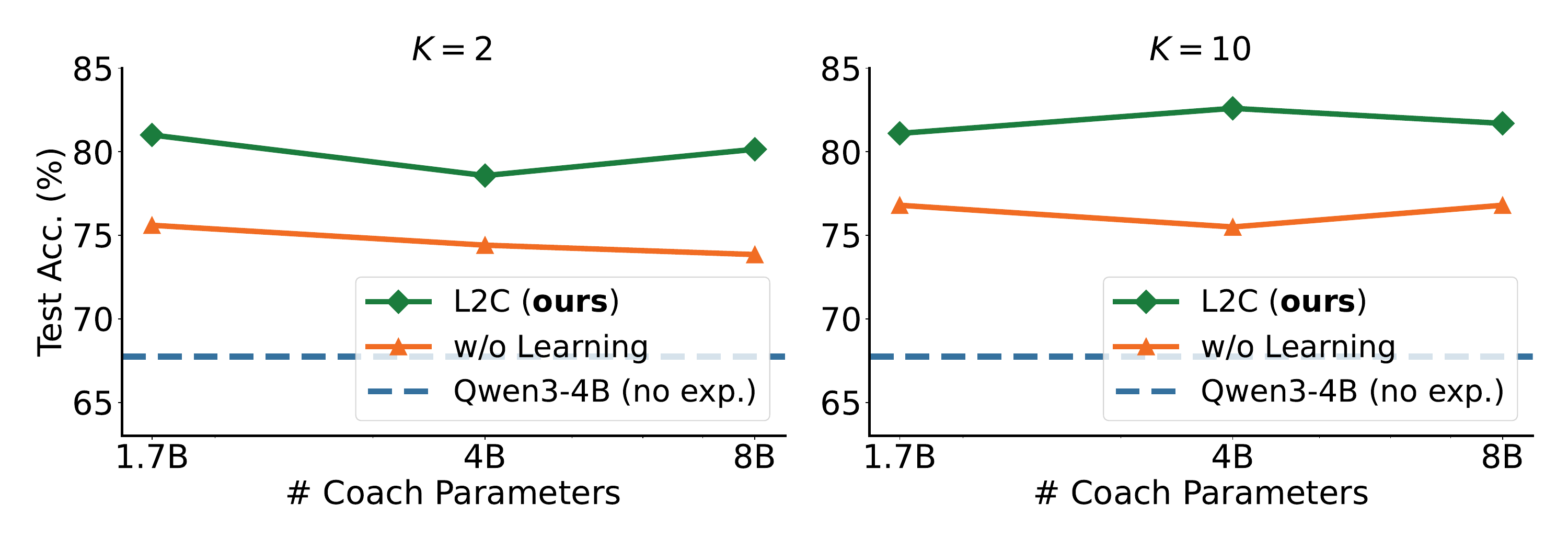}
\caption{Effect of LLM-as-a-Coach size on guided accuracy, with the actor fixed
at Qwen3-4B on DAPO test. \textbf{Left}: $K{=}2$.
\textbf{Right}: $K{=}10$. L2C training consistently improves over
the untrained coach across all LLM-as-a-Coach sizes, yet LLM-as-a-Coach size itself
has little effect on final accuracy under both settings.}
\label{fig:exp-scaling}
\end{figure}
\section{Effect of Model Size}
\label{sec:model-size}

We isolate the effect of LLM-as-a-Coach capacity by fixing the actor at Qwen3-4B
and varying the LLM-as-a-Coach across Qwen3-1.7B, Qwen3-4B and Qwen3-8B, each
trained for $100$ GRPO steps with the same-instance reward (\Eqref{eq:same-reward}). \Cref{fig:exp-scaling} reports both the
$K{=}2$ and the $K{=}10$ settings
on DAPO test.

L2C training consistently improves over the untrained coach across all
LLM-as-a-Coach sizes in both the $K{=}2$ and $K{=}10$
settings, confirming that the training recipe is robust to LLM-as-a-Coach
capacity. However, LLM-as-a-Coach size itself has a surprisingly small effect on
final accuracy: at both $\mathrm{Acc}^{\text{same}}_2$ and $\mathrm{Acc}^{\text{same}}_{10}$, all three
trained LLM-as-a-Coach models reach comparable accuracy, and the untrained variants
likewise cluster within a narrow range.

Actor capacity, by contrast, is the dominant factor. Pairing the same
LLM-as-a-Coach with a stronger actor yields far larger gains than scaling the
LLM-as-a-Coach itself (\Cref{tab:kiter-train}). This suggests that guided
accuracy is primarily rate-limited by the actor's ability to leverage
experiential knowledge, rather than by who produced it. The finding highlights the
practical potential of using a small, lightweight model as the LLM-as-a-Coach to
enhance a much larger actor at modest additional inference cost.

\end{document}